\documentclass[10pt,twocolumn,letterpaper]{article}

\usepackage{iccv}
\usepackage{times}
\usepackage{epsfig}
\usepackage{epstopdf}
\usepackage{graphicx}
\usepackage{amsmath}
\usepackage{amssymb}
\usepackage{gensymb}
\usepackage{multirow}
\usepackage{subfig,caption}
\usepackage{epigraph}
\usepackage{appendix}
\usepackage{comment}
\usepackage{color}
\usepackage{tabu}
\usepackage{xcolor}

\newcommand{\kitti}{KITTI}
\newcommand{\mRed}[1]{\textcolor{black} {#1}}

\iccvfinalcopy 

\def\iccvPaperID{656} 

\begin{document}

\title{Learning to See by Moving}

\author{Pulkit Agrawal\\
UC Berkeley\\
{\tt\small pulkitag@eecs.berkeley.edu}
\and
Jo\~{a}o Carreira \\
UC Berkeley\\
{\tt\small carreira@eecs.berkeley.edu}
\and
Jitendra Malik \\
UC Berkeley\\
{\tt\small malik@eecs.berkeley.edu}
}

\maketitle

    \begin{abstract}
The current dominant paradigm for feature learning in computer vision relies on training neural networks for the task of object recognition using millions of hand labelled images. Is it also possible to learn useful features for a diverse set of visual tasks using any other form of supervision ? In biology, living organisms developed the ability of visual perception for the purpose of moving and acting in the world. Drawing inspiration from this observation, in this work we investigate if the awareness of egomotion can be used as a supervisory signal for feature learning. As opposed to the knowledge of class labels, information about egomotion is freely available to mobile agents. We show that using the same number of training images, features learnt using egomotion as supervision compare favourably to features learnt using class-label as supervision on the tasks of scene recognition, object recognition, visual odometry and keypoint matching. 
\end{abstract}
    \epigraph{"We move in order to see and we see in order to move"}{\textit{J.J Gibson}}

\section{Introduction}
\label{sec:intro}
   
Recent advances in computer vision have shown that visual features learnt by neural networks trained for the task of object recognition using more than a million labelled images are useful for many computer vision tasks like semantic segmentation, object detection and action classification \cite{krizhevsky2012imagenet, donahue2013decaf, agrawal2014analyzing, simonyan2014two}. However, object recognition is one among many tasks for which vision is used. For example, humans use visual perception for recognizing objects, understanding spatial layouts of scenes and performing actions such as moving around in the world. Is there something special about the task of object recognition or is it the case that useful visual representations can be learnt through other modes of supervision? Clearly, biological agents perform complex visual tasks and it is unlikely that they require external supervision in form of millions of labelled examples. Unlabelled visual data is freely available and in theory this data can be used to learn useful visual representations. However, until now  unsupervised learning approaches \cite{bengio2012unsupervised, lee2009convolutional, ranzato2007unsupervised, salakhutdinov2009deep} have not yet delivered on their promise and are nowhere to be seen in current applications on complex real world imagery. 
    
Biological agents use perceptual systems for obtaining sensory information about their environment that enables them to act and accomplish their goals \cite{gibson2013ecological, cutting1986perception}. Both biological and robotic agents employ their motor system for executing actions in their environment. Is it also possible that these agents can use their own motor system as a source of supervision for learning useful perceptual representations? Motor theories of perception have a long history \cite{gibson2013ecological, cutting1986perception}, but there has been little work in formulating computational models of perception that make use of motor information. In this work we focus on visual perception and present a model based on egomotion (i.e. self motion) for learning useful visual representations. When we say useful visual representations \cite{soatto2014visual}, we mean representations that possess the following two characteristics - (1) ability to perform multiple visual tasks and (2) ability of performing new visual tasks by learning from only a few labeled examples provided by an extrinsic teacher.

Mobile agents are naturally aware of their egomotion (i.e. self-motion) through their own motor system. In other words, knowledge of egomotion is ``freely" available. For example, the vestibular system provides the sense of orientation in many mammals. In humans and other animals, the brain has access to information about eye movements and the actions performed by the animal \cite{cutting1986perception}. A mobile robotic agent can estimate its egomotion either from the motor commands it issues to move or from odometry sensors such as gyroscopes and accelerometers mounted on the agent itself. 



We propose that useful visual representations can be learnt by performing the simple task of correlating visual stimuli with egomotion. A mobile agent can be treated like a camera moving in the world and thus the knowledge of egomotion is the same as the knowledge of camera motion. Using this insight, we pose the problem of correlating visual stimuli with egomotion as the problem of predicting the camera transformation from the consequent pairs of images that the agent receives while it moves. Intuitively, the task of predicting camera transformation between two images should force the agent to learn features that are adept at identifying visual elements that are present in both the images (i.e. visual correspondence). In the past, features such as SIFT, that were hand engineered for finding correspondences were also found to be very useful for tasks such as object recognition \cite{lowe1999object,lazebnik2006beyond}. This suggests that egomotion based learning can also result in features that are useful for such tasks.

In order to test our hypothesis of feature learning using egomotion, we trained multilayer neural networks to predict the camera transformation between pairs of images. As a proof of concept, we first demonstrate the usefulness of our approach on the MNIST dataset \cite{lecun1998gradient}. We show that features learnt using our method outperform previous approaches of unsupervised feature learning when class-label supervision is available only for a limited number of examples (section \ref{sub:mnist}) Next, we evaluated the efficacy of our approach on real world imagery. For this purpose, we used image and odometry data recorded from a car moving through urban scenes, made available as part of the KITTI \cite{Geiger2013IJRR} and the San Francisco (SF) city \cite{chen2011city} datasets. This data mimics the scenario of a robotic agent moving around in the world. The quality of features learnt from this data were evaluated on four tasks (1) Scene recognition on SUN \cite{xiao2010sun} (section \ref{sub:sun}), (2) Visual odometery (section \ref{sub:odometry}), (3) Keypoint matching (section \ref{sub:keypoint}) and (4) Object recognition on Imagenet \cite{ILSVRC15} (section \ref{sub:imagenet}). Our results show that for the same amount of training data, features learnt using egomotion as supervision compare favorably to features learnt using class-label as supervision. \mRed{We also show that egomotion based pretraining outperforms a previous approach based on slow feature analysis for unsupervised learning from videos \cite{wiskott2002slow, goroshin2015unsupervised, mobahi2009deep}.} To the best of our knowledge, this work  provides the first effective demonstration of learning visual representations from non-visual access to egomotion information in real world setting.

The rest of this paper is organized as following: In section \ref{sec:other} we discuss the related work, in section \ref{sec:method}, \ref{sub:data}, \ref{sec:results} we present the method, dataset details and we conclude with the discussion in section \ref{sec:discuss}.

\section{Related Work}
\label{sec:other}
Past work in unsupervised learning has been dominated by approaches that pose feature learning as the problem of discovering compact and rich representations of images that are also sufficient to reconstruct the images \cite{bourlard1988auto, barlow1989unsupervised, lee2009convolutional, salakhutdinov2009deep, olshausen1996emergence, ranzato2014video}. Another line of work has focused on learning features that are invariant to transformations either from video \cite{wiskott2002slow, goroshin2015unsupervised, mobahi2009deep} or from images \cite{fischer2014descriptor, ranzato2007unsupervised}. \cite{memisevic2010learning} perform feature learning by modeling spatial transformations using boltzmann machines, but donot evaluate the quality of learnt features.

Despite a lot of work in unsupervised learning (see \cite{bengio2012unsupervised} for a review), a method that works on complex real world imagery is yet to be developed. An alternative to unsupervised learning is to learn features using intrinsic reward signals that are freely available to a system (i.e self-supervised learning). For instance, \cite{hadsell2007online} used intrinsic reward signals available to a robot for learning features that predict path traversability, while \cite{pomerleau1989alvinn} trained neural networks for driving vehicles directly from visual input.

\mRed{In this work we propose to use non-visual access to egomotion information as a form of self-supervision for visual feature learning.} Unlike any other previous work, we show that our method works on real world imagery. Closest to our method is the the work of transforming auto-encoders \cite{hinton2011transforming} that used egomotion to reconstruct the transformed image from an input source image. This work was purely conceptual in nature and the quality of learned features was not evaluated. In contrast, our method uses egomotion as supervision by predicting the transformation between two images using a siamese-like network model \cite{chopra2005learning}.

\mRed{Our method can also be seen as an instance of feature learning from videos. \cite{wiskott2002slow, goroshin2015unsupervised, mobahi2009deep} perform feature learning from videos by imposing the constraint that temporally close frames should have similar feature representations (i.e. slow feature analysis) without accounting for either the camera motion or the motion of objects in the scene. In many settings the camera motion dominates the motion content of the video. Our key observation is that knowledge of camera motion (i.e. egomotion) is \textit{freely} available to mobile agents and can be used as a powerful source of self-supervision.}

\section{A Simple Model of Motion-based Learning}
\label{sec:method}

\begin{figure*}[ht!]
    \begin{center}  
    \includegraphics[width=0.15\linewidth]{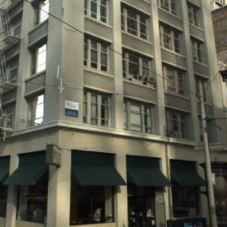} 
    \includegraphics[width=0.15\linewidth]{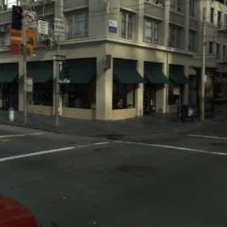} \hspace{1.5mm} 
    \includegraphics[width=0.15\linewidth]{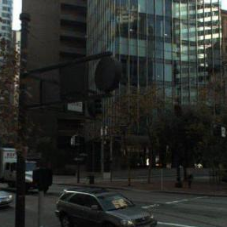}  
    \includegraphics[width=0.15\linewidth]{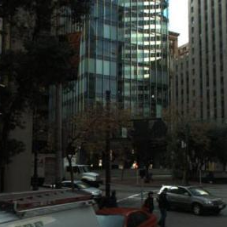} \hspace{1.5mm} 
    \includegraphics[width=0.15\linewidth]{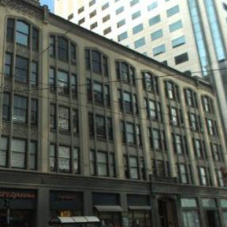} 
    \includegraphics[width=0.15\linewidth]{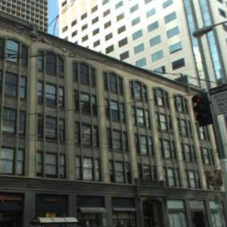} \\ \vspace{2.5mm}
    \includegraphics[width=0.15\linewidth]{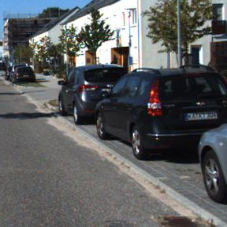} 
    \includegraphics[width=0.15\linewidth]{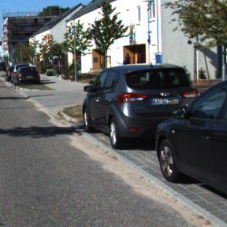} \hspace{1.5mm} 
    \includegraphics[width=0.15\linewidth]{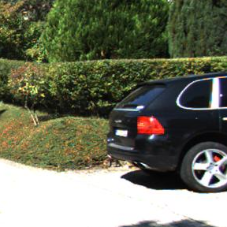}  
    \includegraphics[width=0.15\linewidth]{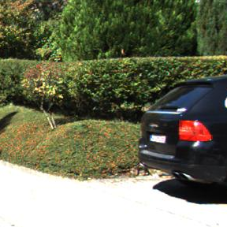} \hspace{1.5mm} 
    \includegraphics[width=0.15\linewidth]{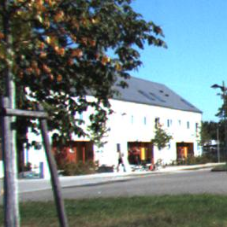} 
    \includegraphics[width=0.15\linewidth]{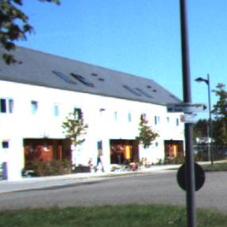} \\
    \end{center}
    \caption{Exploring the utility of egomotion as supervision for learning useful visual features. A mobile agent equipped with visual sensors receives a sequence of images as inputs while it moves in its environment. The movement of the agent is equivalent to the movement of a camera. In this work, egomotion based learning is posed as the problem of predicting camera transformation from image pairs. The top and bottom rows of the figure show some sample image pairs from the SF and KITTI datasets that were used for feature learning.}
    \label{fig:sample}
\end{figure*}

We model the visual system of the agent  with a Convolutional Neural Network (CNN, \cite{lecun1989backpropagation}). The agent optimizes its visual representations (i.e. updating the weights of the CNN) by minimizing the error between the egomotion information (i.e. camera transformation) obtained from its motor system and egomotion predicted using its visual inputs only. Performing this task is equivalent to training a CNN with two streams (i.e. Siamese Style CNN or SCNN\cite{chopra2005learning}) that takes two images as inputs and predicts the egomotion that the agent underwent as it moved between the two spatial locations from which the two images were obtained. In order to learn useful visual representations, the agent continuously performs this task as it moves around in its environment.

In this work we use the pretraining-finetuning paradigm for evaluating the utility of learnt features. Pretraining is the process of optimizing the weights of a randomly initialized CNN for an auxiliary task that is not the same as the target task. Finetuning is the process of modifying the weights of a pretrained CNN for the given target task. \mRed{Our experiments compare the utility of features learnt using egomotion based pretraining against class-label based and slow-feature based pretraining on multiple target tasks.}


\subsection{\mRed{Two Stream Architecture}}
\label{two-stream}
\begin{figure}[t]
    \begin{center}  
    \includegraphics[width=0.9\linewidth]{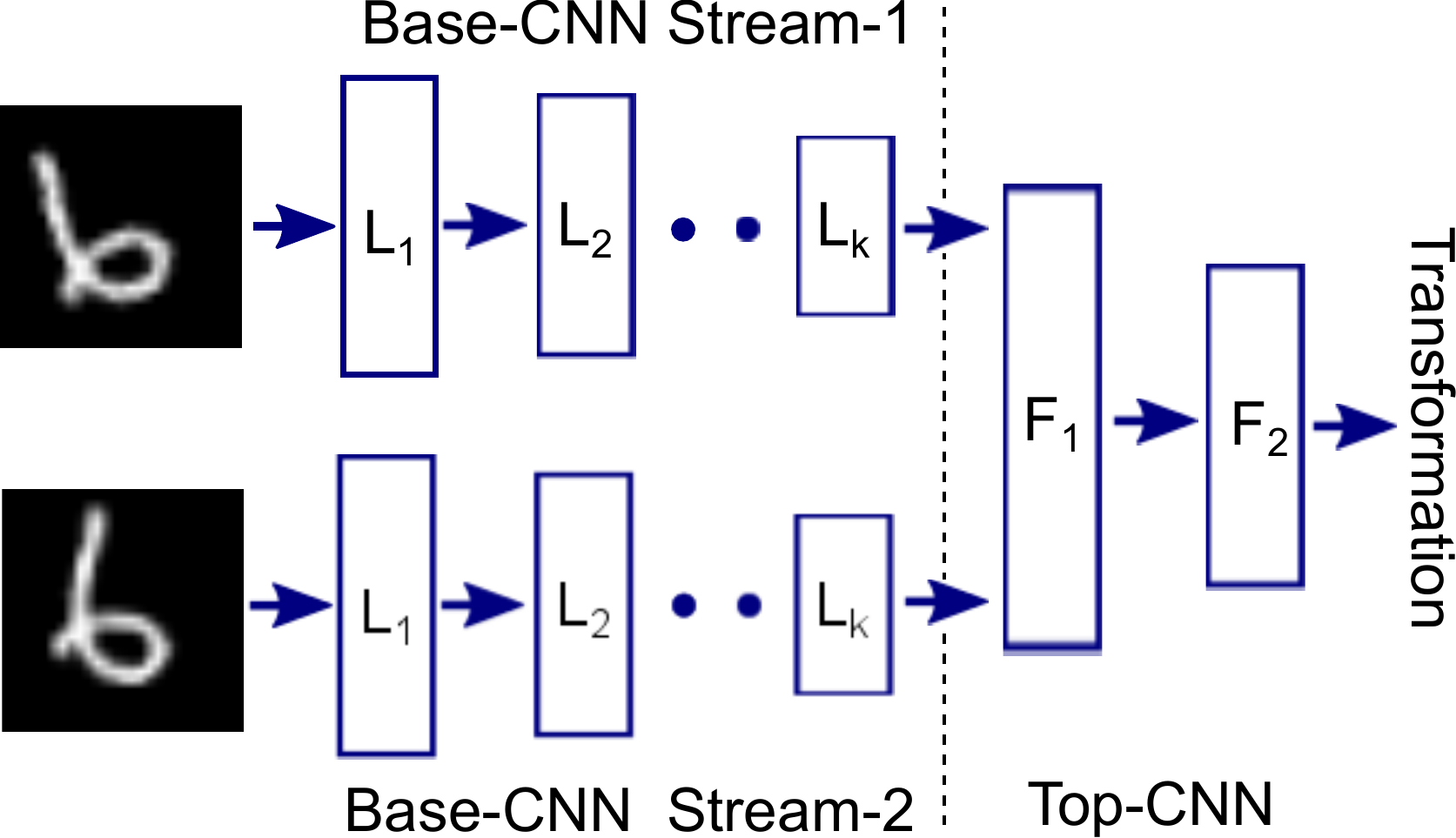} \hspace{5mm} 
    \end{center}
    \caption{Description of the method for feature learning. Visual features are learnt by training a Siamese style Convolutional Neural Network (SCNN, \cite{chopra2005learning}) that takes as inputs two images and predicts the transformation between the images (i.e. egomotion). Each stream of the SCNN (called as Base-CNN or BCNN) computes features for one image. The outputs of two BCNNs are concatenated and passed as inputs to a second multilayer CNN called as the Top-CNN (TCNN) (shown as layers $F_1$, $F_2$).  The two BCNNs have the same architecture and share weights. After feature learning, TCNN is discarded and a single BCNN stream is used as a standard CNN for extracting features for performing target tasks like scene recognition.}
    \label{fig:architecture}
\end{figure}

Each stream of the CNN independently computes features for one image. Both streams share the same architecture and the same set of weights and consequently perform the same set of operations for computing features. The individual streams have been called as Base-CNN (BCNN). Features from two BCNNs are concatenated and passed downstream into another CNN called as the Top-CNN (TCNN) (see figure \ref{fig:architecture}). TCNN is responsible for using the BCNN features to predict the camera transformation between the input pair of images. After pretraining, the TCNN is removed and a single BCNN is used as a standard CNN for feature computation for the target task.


\subsection{\mRed{Shorthand for CNN architectures}}
The abbreviations Ck, Fk, P, D, Op represent a convolutional(C) layer with k filters, a fully-connected(F) layer with k filters, pooling(P), dropout(D) and the output(Op) layers respectively. We used ReLU non-linearity after every convolutional/fully-connected layer, except for the output layer. The dropout layer was always used with dropout of 0.5. The output layer was a fully connected layer with number of units equal to the number of desired outputs. As an example of our notation, C96-P-F500-D refers to a network with 96 filters in the convolution layer followed by ReLU non-linearity, a pooling layer, a fully-connected layer with 500 unit, ReLU non-linearity and a dropout layer. We used \cite{caffe} for training all our models. 

\subsection{\mRed{Slow Feature Analysis (SFA) Baseline}}
\mRed{Slow Feature Analysis (SFA) is a method for feature learning based on the principle that useful features change slowly in time. 
We used the following contrastive loss formulation of SFA \cite{chopra2005learning, mobahi2009deep}}, \\ \\
\noindent $L(x_{t_1}, x_{t_2}, W) = $
\begin{equation}
\label{eq:sfa}
\begin{cases} 
    D(x_{t_1}, x_{t_2}) & \text{if } |t_1 - t_2| \leq T \\
    1 - max \big(0, m - D(x_{t_1}, x_{t_2}) \big)  & \text{if } |t_1 - t_2| > T
\end{cases}
\end{equation}

\mRed{
where, $L$ is the loss, $x_{t_1}, x_{t_2}$ refer to feature representations of frames observed at times $t_1, t_2$ respectively, $W$ are the parameters that specify the feature extraction process, $D$ is a measure of distance with parameter, $m$ is a predefined margin and $T$ is a predefined time threshold for determining whether the two frames are temporally close or not. In this work, $x_t$ are features computed using a CNN with weights $W$ and $D$ was chosen to be the L2 distance. SFA pretraining was performed using two stream architectures that took pairs of images as inputs and produced outputs $x_{t_1}, x_{t_2}$ as outputs from the two streams respectively.} 

\subsection{Proof of Concept using MNIST}
\label{sub:mnist}
On MNIST, egomotion was emulated by generating synthetic data consisting of random transformation (translations and rotations) of digit images. From the training set of 60K images, digits were randomly sampled and then transformed using two different sets of random transformations to generate image pairs. CNNs were trained for predicting the transformations between these image pairs. 

\subsubsection{Data}
\label{sub:mnist-data}
For egomotion based pretraining, relative translation between the digits was constrained to be an integer value in the range [-3, 3] and relative rotation $\theta$ was constrained to lie within the range [-30\degree, 30\degree]. The prediction of transformation was posed as a classification task with three separate soft-max losses (one each for translation along X, Y axes and the rotation about Z-axis). SCNN was trained to minimize the sum of these three losses. Translations along X, Y were separately binned into seven uniformly spaced bins each. The rotations were binned into bins of size 3\degree each resulting into a total of 20 bins (or classes). 
\mRed{For SFA based pretraining, image pairs with relative translation in the range [-1, 1] and relative rotation within [-3\degree, 3\degree] were considered to be temporally close to each other (see equation \ref{eq:sfa}).} A total of 5 million image pairs were used for both pretraining procedures.

\subsubsection{Network Architectures}
\label{sub:mnist-arch}
We experimented with multiple BCNN architectures \mRed{and chose the optimal architecture for each pretraining method separately.} For egmotion based pretraining, the two BCNN streams were concatenated using the TCNN: \textit{F1000-D-Op}. Pretraining was performed for 40K iterations (i.e. 5M examples) using an initial learning rate of 0.01 which was reduced by a factor of 2 after every 10K iterations.

The following architecture was used for finetuning: \textit{BCNN-F500-D-Op}. In order to evaluate the quality of BCNN features, the learning rate of all layers in the BCNN were set to 0 during finetuning for digit classification. Finetuning was performed for 4K iterations (which is equivalent to training for 50 epochs for the 10K labelled training examples) with a constant learning rate of 0.01.

\subsubsection{Results}
\label{sub:mnist-res}
The BCNN features were evaluated by computing the error rates on the task of digit classification using 100, 300, 1K and 10K class-labelled examples for training. These sets were constructed by randomly sampling digits from the standard training set of 60K digits. For this part of the experiment, the original digit images were used (i.e. without any transformations or data augmentation). The standard test set of 10K digits was used for evaluation and error rates averaged across 3 runs are reported in table \ref{table:mnist}.

\mRed{The BCNN architecture: C96-P-C256-P, was found to be optimal for egomotion and SFA based pretraining and also for training from scratch (i.e. random weight initialization). Results for other architectures are provided in the supplementary material. For SFA based pretraining, we experimented with multiple values of the margin $m$ and found that $m=10, 100$ led to the best performance.} Our method outperforms convolutional deep belief networks \cite{lee2009convolutional}, a previous approach based on learning features invariant to transformations \cite{ranzato2007unsupervised} and \mRed{SFA based pretraining.}   

\setlength{\tabcolsep}{2pt}
\begin{table}[t!]
    \begin{center}
    \caption{Comparison of various pretraining methods on MNIST reveals that egomotion based pretraining outperforms many previous approaches for unsupervised learning. The performance is reported as the \textbf{error rate}.}
    \vspace{0.3em}
    \label{table:mnist}
    \scalebox{1.0}{
    \begin{tabu}{l|cccc}
   Method & \multicolumn{4}{c}{\# examples for finetuning} \\
    \hline
     &  \textbf{100} & \textbf{300} & \textbf{1000} & \textbf{10000} \\
    \hline
    Autoencoder \cite{caffe} & 24.1 &   12.2     &     7.7        &    4.8    \\
    Ranzato et al. \cite{ranzato2007unsupervised} &     -        &    7.18      &     3.21      &     \textbf{0.85}   \\
    Lee et al. \cite{lee2009convolutional}      &     -        &      -       &     2.62      &     -      \\
    \hline
    \hline
    Train from Scratch     & 20.1 & 8.3 & 4.5 & 1.6 \\ 
   SFA (m=10) & 11.2 & 6.4 & 3.5 & 2.1 \\ 
   SFA (m=100)& 11.9 & 6.4 & 4.8 & 4.7 \\ 
    \hline
    \hline
    Egomotion (ours) & \textbf{8.7} & \textbf{3.6} & \textbf{2.0} & 0.9 \\
    \hline
    \end{tabu}}
    \end{center}
\end{table}
\setlength{\tabcolsep}{1.4pt}

\section{Learning Visual Features From Egomotion in Natural Environments}
\label{sub:data}

We used two main sources of real world data for feature learning: the KITTI and SF datasets, which were collected using cameras and  odometry sensors mounted on a car driving through urban scenes. Details about the data, the experimental procedure, the network architectures and the results are provided in sections \ref{sub:kitti}, \ref{sub:sf}, \ref{sub:net-real} and \ref{sec:results} respectively.

\subsection{\kitti  ~Dataset}
\label{sub:kitti}
The \kitti ~dataset provided odometry and image data recorded during 11 short trips of variable length made by a car moving through urban landscapes. The total number of frames in the entire dataset was 23,201. Out of 11, 9 sequences were used for training and 2 for validation. The total number of images in the training set was 20,501. 

\mRed{The odometry data was used to compute the camera transformation between pairs of images recorded from the car. The direction in which the camera pointed was assumed to be the Z axis and the image plane was taken to be the XY plane. X-axis and Y-axis refer to horizontal and vertical directions in the image plane.} As significant camera transformations in the KITTI data were either due to translations along the Z/X axis or rotation about the Y axis, only these three dimensions were used to express the camera transformation. The rotation was represented as the euler angle about the Y-axis. The task of predicting the transformation between pair of images was posed as a classification problem. The three dimensions of camera transformation were individually binned into 20 uniformly spaced bins each. The training image pairs were selected from frames that were at most $\pm 7$ frames apart to ensure that images in any given pair would have a reasonable overlap. \mRed{For SFA based pretraining, pairs of frames that were separated by atmost $\pm 7$ frames were considered to be temporally close to each other.}

The SCNN was trained to predict camera transformation from pairs of $ 227 \times 227$ pixel sized image regions extracted from images of overall size $ 370 \times 1226 $ pixels. For each image pair, the coordinates for cropping image regions were randomly chosen. Figure \ref{fig:sample} illustrates typical image crops.

\subsection {SF Dataset}
\label{sub:sf}
SF dataset provides camera transformation between $\approx$ 136K pairs of images (constructed from a set of 17,357 unique images). This dataset was constructed using Google StreetView \cite{chen2011city}. $ \approx 130K$ image pairs were used for training and $ \approx 6K$ pairs for validation. 

Just like KITTI, the task of predicting camera transformation was posed as a classification problem. Unlike KITTI, significant camera transformation was found along all six dimensions of transformation (i.e. the 3 euler angles and the 3 translations).  Since, it is unreasonable to expect that visual features can be used to infer big camera transformations, rotations between [-30\degree, 30\degree] were binned into 10 uniformly spaced bins and two extra bins were used for rotations larger and smaller than 30\degree and -30\degree respectively. The three translations were individually binned into 10 uniformly spaced bins each. Images were resized to a size of $360 \times 480$ and image regions of size $227 \times 227 $ were used for training the SCNN.

\subsection{Network Architecture}
\label{sub:net-real}
BCNN closely followed the architecture of first five AlexNet layers \cite{krizhevsky2012imagenet}: \textit{C96-P-C256-P-C384-C384-C256-P}. TCNN architecture was: \textit{C256-C128-F500-D-Op}. The convolutional filters in the TCNN were of spatial size $3 \times 3$. The networks were trained for 60K iterations with a batch size of 128. The initial learning rate was set to 0.001 and was reduced by a factor of two after every 20K iterations.

\begin{figure}[t]
       \begin{center}
        \subfloat[KITTI-Net]{\includegraphics[width=0.40\linewidth]{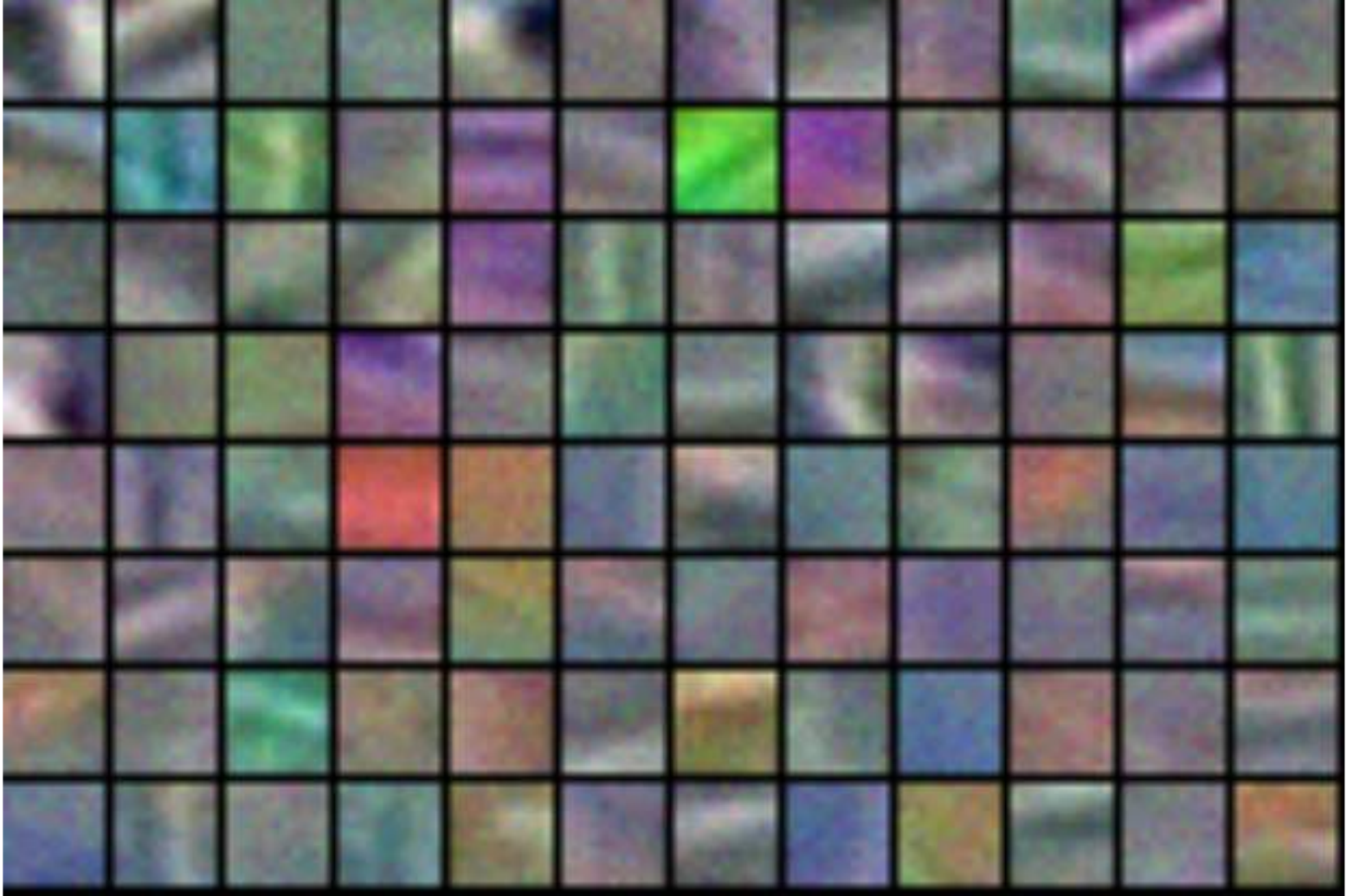}} \hspace{2mm}
        \subfloat[SF-Net]{\includegraphics[width=0.40\linewidth]{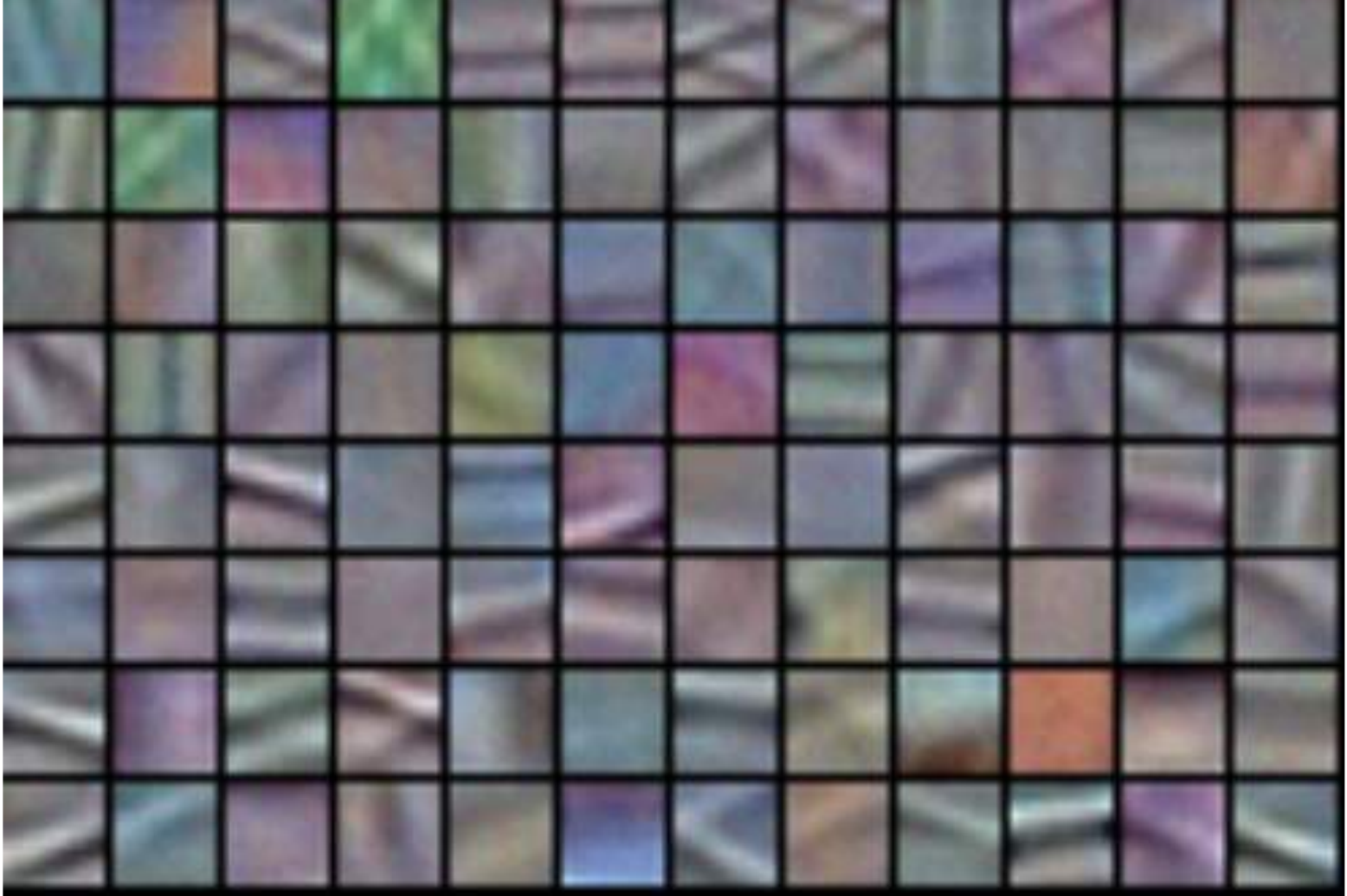}}
       \end{center}
       \caption{Visualization of layer 1 filters learnt by egomotion based pretraining on (a) KITTI and (b) SF datasets. A large majority of layer-1 filters are color detectors and some of them are edge detectors. This is expected as color is a useful cue for determining correspondences between image pairs.}
    \label{fig:filters}
\end{figure}

We term the networks pretrained \mRed{using egomotion on KITTI and SF datasets as KITTI-Net and SF-Net respectively.} \mRed{The net pretrained on KITTI with SFA is called KITTI-SFA-Net}. Figure \ref{fig:filters} shows the layer-1 filters of KITTI-Net and SF-Net. A large majority of layer-1 filters are color detectors, while some of them are edge detectors. As color is a useful cue for determining correspondences between closeby frames of a video sequence, learning of color detectors as layer-1 filters is not surprising. The fraction of filters that detect edges is higher for the SF-Net. This is not surprising either, because higher fraction of images in the SF dataset contain structured objects like buildings and cars.

\section{Evaluating Motion-based Learning}
\label{sec:results}

For evaluating the merits of the proposed approach, features learned using egomotion based supervision were compared against features learned using class-label and SFA based supervision on the challenging tasks of scene recognition, intra-class keypoint matching and visual odometry and object recognition. The ultimate goal of feature learning is to find features that can generalize from only a few supervised examples on a new task. Therefore it makes sense to evaluate the quality of features when only a few labelled examples for the target task are provided. Consequently, the scene and object recognition experiments were performed in the setting when only 1-20 labelled examples per class were available for finetuning.  

The KITTI-Net and SF-Net (examples of models trained using egomotion based supervision) were trained using only only $\approx$ 20K unique images. To make a fair comparison with class-label based supervision, a model with AlexNet architecture was trained using only 20K images taken from the training set of ILSVRC12 challenge (i.e. 20 examples per class). This model has been referred to as AlexNet-20K. In addition, some experiments presented in this work also make comparison with AlexNet models trained with 100K and 1M images that have been named as AlexNet-100K and AlexNet-1M respectively.



\subsection{Scene Recognition}
\label{sub:sun}
SUN dataset consisting of 397 indoor/outdoor scene categories was used for evaluating scene recognition performance. This dataset provides 10 standard splits of 5 and 20 training images per class and a standard test set of 50 images per class. Due to time limitation of running 10 runs of the experiment, we evaluated the performance using only 3 train/test splits.

For evaluating the utility of CNN features produced by different layers, separate linear (SoftMax) classifiers were trained on features produced by individual CNN layers (i.e. BCNN layers of KITTI-Net, KITTI-SFA-Net and SF-Net). Table \ref{table:sun} reports recognition accuracy (averaged over 3 train/test splits) for various networks considered in this study. KITTI-Net outperforms SF-Net and is comparable to AlexNet-20K. This indicates that given a fixed budget of pretraining images, egomotion based supervision learns features that are almost as good as the features using class-based supervision on the task of scene recognition. \mRed{The performance of features computed by layers 1-3 (abbreviated as L1, L2, L3 in table \ref{table:sun}) of the KITTI-SFA-Net and KITTI-Net is comparable, whereas layer 4, 5 features of KITTI-Net significantly outperform layer 4, 5 features of KITTI-SFA-Net. This indicates that egomotion based pretraining results into learning of higher-level features, while SFA based pretraining results into learning of lower-level features only.}    

The KITTI-Net outperforms GIST\cite{oliva2006building}, which was specifically developed for scene classification, but is outperformed by Dense SIFT with spatial pyramid matching (SPM) kernel \cite{lazebnik2006beyond}. The KITTI-Net was trained using limited visual data ($ \approx 20K frames$) containing visual imagery of limited diversity. The KITTI data mainly contains images of roads, buildings, cars, few pedestrians, trees and some vegetation. It is in fact surprising that a network trained on data with such little diversity is competitive on classifying indoor and outdoor scenes with the AlexNet-20K that was trained on a much more diverse set of images. We believe that with more diverse training data for egomotion based learning, the performance of learnt features will be better than currently reported numbers.

The KITTI-Net outperformed the SF-Net except for the performance of layer 1 (L1). As it was possible to extract a larger number of image region pairs from the KITTI dataset as compared to the SF dataset (see section \ref{sub:kitti}, \ref{sub:sf}), the result that KITTI-Net outperforms SF-Net is not surprising. Because KITTI-Net was found to be superior to the SF-Net in this experiment, the KITTI-Net was used for all other experiments described in this paper.  

\setlength{\tabcolsep}{2pt}
    \begin{table*}[t!]
    \begin{center}
    \caption{Comparing the \textbf{accuracy} of neural networks pre-trained using motion-based and class-label based supervision for the task of scene recognition on the SUN dataset. The performance of layers 1-6 (labelled as L1-L6) of these networks was evaluated after finetuning the network using 5/20 images per class from the SUN dataset. The performance of the KITTI-Net (i.e. motion-based pretraining) fares favorably with a network pretrained on Imagenet (i.e. class-based pretraining) with the same number of pretraining images (i.e. 20K).}
    \vspace{0.3em}
    \label{table:sun}
    \scalebox{0.9}{
    \begin{tabu}{c|c|c|c|c|c|c|c|c|c|c|c|c|c|c|c|c}
    \textbf{Method} & \textbf{Pretrain Supervision}  & \textbf{\#Pretrain} & \textbf{\#Finetune} &  \textbf{L1} & \textbf{L2} & \textbf{L3} & \textbf{L4} & \textbf{L5} & \textbf{L6} &  \textbf{\#Finetune} &\textbf{L1} & \textbf{L2} & \textbf{L3} & \textbf{L4} & \textbf{L5} & \textbf{L6}  \\
    \hline
    AlexNet-1M  & \multirow{2}{*}{Class-Label} & 1M  & 5  &  5.3 &10.5& 12.1 & 12.5 &18.0 &  23.6 & 20 & 11.8 &22.2& 25.0 & 26.8 & 33.3 & 37.6   \\
    \cline{3-17}
    AlexNet-20K &  & 20K & 5 & 4.9  & 6.3 & 6.6 & 6.3 & 6.6 & 6.7 &  20 & 8.7  &12.6& 12.4 & 11.9 & 12.5 & 12.4   \\
    \hline
    \hline
   KITTI-SFA-Net & Slowness & 20.5K & 5 & 4.5  &  5.7  & 6.2  &  3.4   &  0.5    &   -  & 20  &  8.2   &  11.2  & 12.0  &  7.3  &  1.1  & - \\
    \hline
    \hline
    SF-Net        &  \multirow{2}{*}{Egomotion} &   18K & 5  &  4.4 & 5.2 & 4.9 & 5.1  & 4.7   & -     & 20 &  8.6 &11.6 & 10.9 &10.4 & 9.1   & -    \\
     \cline{3-17}
    KITTI-Net    &  &   20.5K & 5  & 4.3  &6.0  & 5.9 & 5.8  & 6.4  & - & 20 & 7.9  &12.2 &12.1 & 11.7 & 12.4 & -    \\
    \hline
    \hline
    GIST \cite{xiao2010sun}         & Human  & -    & 5  &  \multicolumn{6}{|c|}{6.2} & 20 &  \multicolumn{6}{|c}{11.6} \\ 
    \hline
    SPM \cite{xiao2010sun}          & Human  & -    & 5  &  \multicolumn{6}{|c|}{8.4}  & 20 &  \multicolumn{6}{|c}{16.0} \\
    \hline
    \end{tabu}}
    \end{center}
    \end{table*}
    \setlength{\tabcolsep}{1.4pt}

\subsection{Object Recognition}
\label{sub:imagenet}
If egomotion based pretraining learns useful features for object recognition, then a net initialized with KITTI-Net weights should outperform a net initialized with random weights on the task of object recognition. For testing this, we trained CNNs using 1, 5, 10 and 20 images per class from the ILSVRC-2012 challenge. As this dataset contains 1000 classes, the total number of training examples available for training for these networks were 1K, 5K, 10K and 20K respectively. \mRed{All layers of KITTI-Net, KITTI-SFA-Net and AlexNet-Scratch (i.e. CNN with random weight initialization) were finetuned for image classification.} 

The results of the experiment presented in table \ref{table:imagenet} show that \mRed{egomotion based supervision (KITTI-Net) clearly outperforms SFA based supervision(KITTI-SFA-Net) and AlexNet-Scratch}. As expected, the improvement offered by motion-based pretraining is larger when the number of examples provided for the target task are fewer. These result show that egomotion based pretraining learns features useful for object recognition.

\setlength{\tabcolsep}{2pt}
\begin{table}[t!]
    \begin{center}
    \caption{Top-5 \textbf{accuracy} on the task of object recognition on the ILSVRC-12 validation set. AlexNet-Scratch refers to a net with AlexNet architecture initialized with randomly weights. \mRed{The weights of KITTI-Net and KITTI-SFA-Net were learned using egomotion based and SFA based supervision on the KITTI dataset respectively.} All the networks were finetuned using  $1, 5, 10, 20$ examples per class. \mRed{The KITTI-Net clearly outperforms AlexNet-Scratch and KITTI-SFA-Net.}}
    \vspace{0.3em}
    \label{table:imagenet}
    \begin{tabu}{c|c|c|c|c}
    \textbf{Method}       & 1         & 5         & 10        & 20         \\
    \hline
    AlexNet-Scratch    & 1.1       & 3.1       & 5.9       & 14.1       \\
    \hline
   KITTI-SFA-Net (Slowness) &1.5 & 3.9 & 6.1 &14.9 \\
    \hline
    KITTI-Net (Egomotion)             & \textbf{2.3}          & \textbf{5.1}       & \textbf{8.6}       & \textbf{15.8}       \\ 
    \hline
    \end{tabu}
    \end{center}
\end{table}    
\setlength{\tabcolsep}{1.4pt}

\subsection{Intra-Class Keypoint Matching}
\label{sub:keypoint}
Identifying the same keypoint of an object across different instances of the same object class is an important visual task. Visual features learned using \mRed{egomotion, SFA and class-label based supervision were evaluated for this task using} keypoint annotations on the PASCAL dataset \cite{bourdev2010detecting}. 

Keypoint matching was computed in the following way: First, ground-truth object bounding boxes (GT-BBOX) from PASCAL-VOC2012 dataset were extracted and re-sized (while preserving the aspect ratio) to ensure that the smaller side of the boxes was of length 227 pixels. Next, feature maps from layers 2-5 of various CNNs were computed for every GT-BBOX. The keypoint matching score was computed between all pairs of GT-BBOX belonging to the same object class. For given pair of GT-BBOX, the features associated with keypoints in the first image were used to predict the location of the same keypoints in the second image. The normalized pixel distance between the actual and predicted keypoint locations was taken as the error in keypoint matching. More details about this procedure have been provided in the supp. materials.

\begin{figure*}[t]
       \begin{center}
 \includegraphics[width=0.44\linewidth]{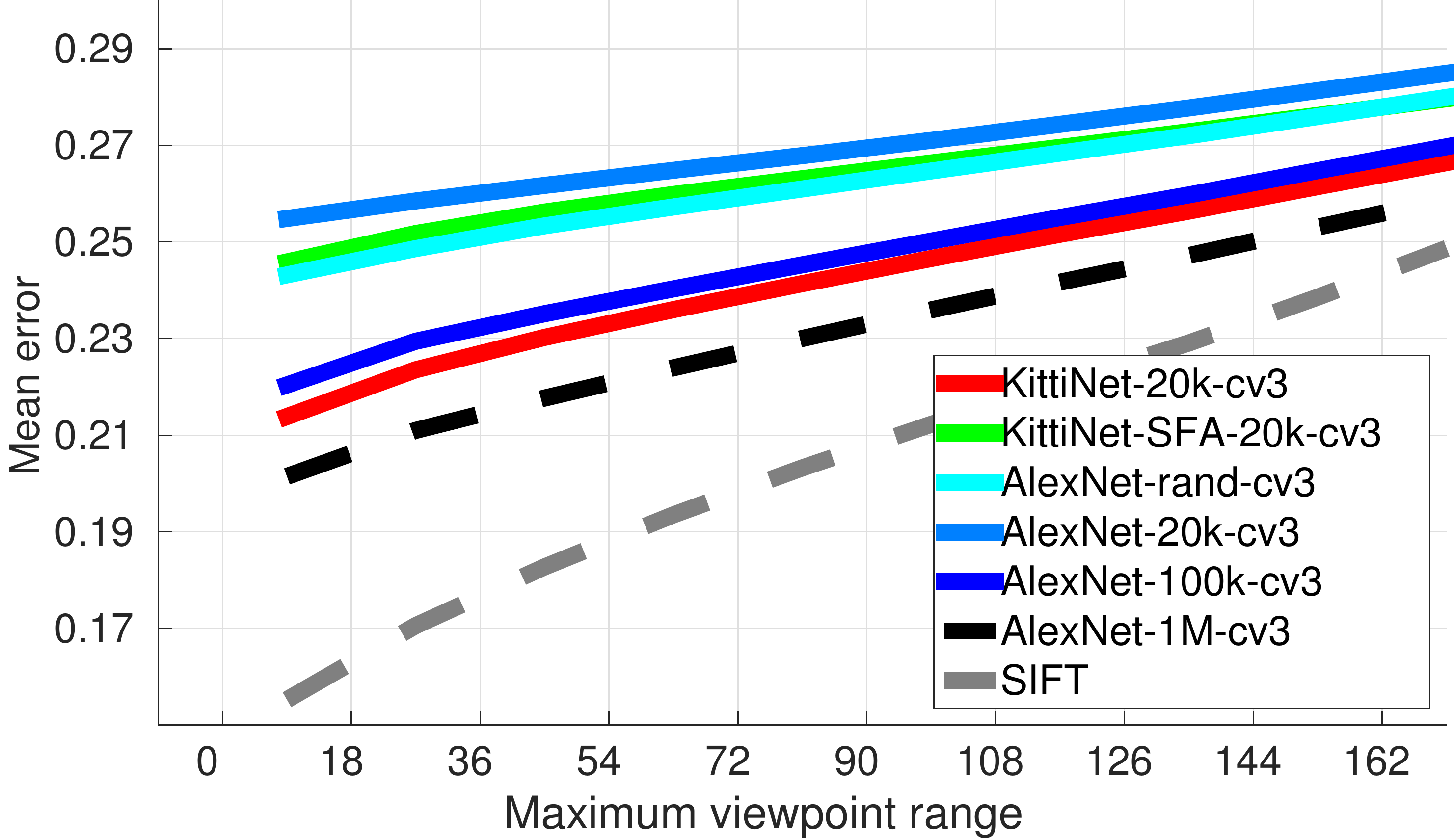}
 \includegraphics[width=0.44\linewidth]{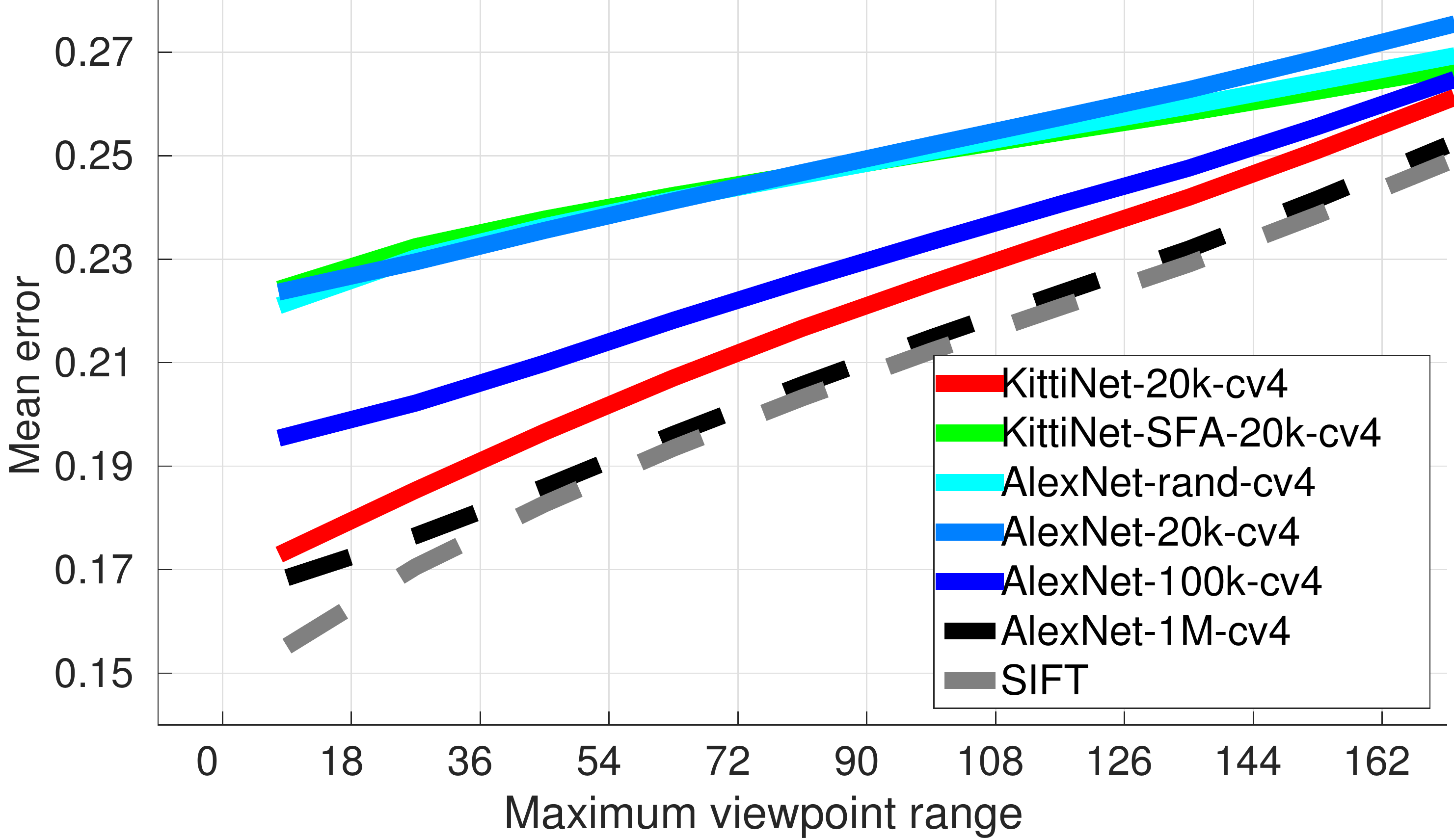}
       \end{center}
       \caption{\mRed{Intra-class keypoint matching error as a function of viewpoint distance averaged over 20 PASCAL objects using features from layers conv3 (left) and conv4 (right) of various CNNs used in this work. Please see the text for more details.}}
    \label{fig:alignmnent_err}
    \end{figure*}

It is natural to expect that accuracy of keypoint matching would depend on the camera transformation between the two viewpoints of the object(i.e. viewpoint distance). In order to make a holistic evaluation of the utility of features learnt by different pretraining methods on this task, matching error was computed as a function of viewpoint distance \cite{vicente2014reconstructing}. Figure \ref{fig:alignmnent_err} reports the matching error averaged across all keypoints, all pairs of GT-BBOX and all classes using features extracted from layers conv-3 and conv-4. 

KITTI-Net trained only with 20K unique frames was superior to AlexNet-20K and AlexNet-100K and inferior only to AlexNet-1M. A net with AlexNet architecture initialized with random weights (AlexNet-Rand), surprisingly performed better than AlexNet-20K. One possible explanation for this observation is that with only 20K examples, features learnt by AlexNet-20K only capture coarse global appearance of objects and are therefore poor at keypoint matching. SIFT has been hand engineered for finding correspondences across images and performs as well as the best AlexNet-1M features for this task (i.e. conv-4 features). \mRed{KITTI-Net also significantly outperforms KITTI-SFA-Net. These results indicate that features learnt by egomotion based pretraining are superior to SFA and class-label based pretraining for the task of keypoint matching.}

\subsection{Visual Odometry}
\label{sub:odometry}
Visual odometry is the task of estimating the camera transformation between image pairs. All layers of KITTI-Net and AlexNet-1M were finetuned for 25K iterations using the training set of SF dataset on the task of visual odometry (see section \ref{sub:sf} for task description). The performance of various CNNs was evaluated on the validation set of SF dataset and the results are reported in table \ref{table:odo}.

\mRed{Performance of KITTI-Net was either superior or comparable to AlexNet-1M on this task. As the evaluation was made on the SF dataset itself, it was not surprising that on some metrics SF-Net outperformed KITTI-Net.} The results of this experiment indicate that egomotion based feature learning is superior to class-label based feature learning on the task of visual odometry.

\setlength{\tabcolsep}{2pt}
\begin{table}[t!]
    \begin{center}
    \caption{Comparing the \textbf{accuracy} of various pretraining methods on the task of visual odometry.}
    \vspace{0.3em}
    \label{table:odo}
    \scalebox{0.90}{
    \begin{tabular}{c|c|c|c|c|c|c}
    \textbf{Method}       & \multicolumn{3}{c|}{\textbf{Translation Acc.}} &  \multicolumn{3}{c}{\textbf{Rotation Acc.}} \\
    \hline
                          &  $\delta X$ & $\delta Y$ & $\delta Z$ & $\delta \theta_1$ & $\delta \theta_2$ & $\delta \theta_3$ \\ 
    \hline
    \mRed{SF-Net}    &  40.2  & \textbf{58.2}   & 38.4       & 45.0 & \textbf{44.8} & 40.5       \\
    \hline
    KITTI-Net             &  \textbf{43.4}  & 57.9   & \textbf{40.2} & \textbf{48.4} & 44.0 & \textbf{41.0} \\ 
    \hline
    AlexNet-1M           &  41.8 & 58.0  & 39.0         & 46.0 & 44.5 & 40.5 \\
    \hline
    \end{tabular}}
    \end{center}
\end{table}    
\setlength{\tabcolsep}{1.4pt}


\section{Discussion}
\label{sec:discuss}
In this work, we have shown that egomotion is a useful source of intrinsic supervision for visual feature learning in mobile agents. In contrast to class labels, knowledge of egomotion is ''freely" available. On MNIST, egomotion-based feature learning outperforms many previous unsupervised methods of feature learning. Given the same budget of pretraining images, on task of scene recognition, egomotion-based learning performs almost as well as class-label-based learning. Further, egomotion based features outperform features learnt by a CNN trained using class-label supervision on two orders of magnitude more data (AlexNet-1M) on the task of visual odometry and one order of magnitude more data on the task of intra-class keypoint matching. In addition to demonstrating the utility of egomotion based supervision, these results also suggest that features learnt by class-label based supervision are not optimal for all visual tasks. This means that future work should look at what kinds of pretraining are useful for what tasks. 
    
One potential criticism of our work is that we have trained and evaluated  high capacity deep models on relatively little data (e.g. only 20K unique images available on the KITTI dataset). In theory, we could have learnt better features by downsizing the networks. For example, in our experiments with MNIST we found that pretraining a 2-layer network instead of 3-layer results in better performance (table \ref{table:mnist}). In this work, we have made a conscious choice of using standard deep models because the main goal of this work was not to explore novel feature extraction architectures but to investigate the value of egmotion for learning visual representations on architectures known to perform well on practical applications. Future research focused on exploring architectures that are better suited for egomotion based learning can only make a stronger case for this line of work. While egomotion is freely available to mobile agents, there are currently no publicly available datasets as large as Imagenet. Consequently, we were unable to evaluate the utility of motion-based supervision across the full spectrum of training set sizes. 
    
In this work, we chose to first pretrain our models using a base task (i.e. egomotion) and then finetune these models for target tasks. An equally interesting setting is that of online learning where the agent has continuous access to intrinsic supervision (such as egomotion) and occasional explicit access to extrinsic teacher signal (such as the class labels). We believe that such a training procedure is likely to result in learning of better features. Our intuition behind this is that seeing different views of the same instance of an object (say) car, may not be sufficient to learn that different instances of the car class should be grouped together. The occasional extrinsic signal about object labels may prove useful for the agent to learn such concepts. 
Also, current work makes use of passively collected egomotion data and it would be interesting to investigate if it is possible to learn better visual representations if the agent can actively decide on how to explores its environment (i.e. active learning \cite{bajcsy1988active}). 

\section*{Acknowledgements}
This  work  was  supported  in  part  by   ONR  MURI-N00014-14-1-0671. 
Pulkit Agrawal was partially supported by Fulbright Science and Technology Fellowship. Jo\~{a}o
Carreira was supported by the Portuguese Science Founda-
tion, FCT, under grant SFRH/BPD/84194/2012.  We grate-
fully acknowledge NVIDIA corporation for the donation of
Tesla GPUs for this research.

\section*{Appendix}
\begin{appendices}
\section{Keypoint Matching Score}
\label{sec:key}
\begin{figure*}
\begin{center}
  \includegraphics[width=0.98\linewidth]{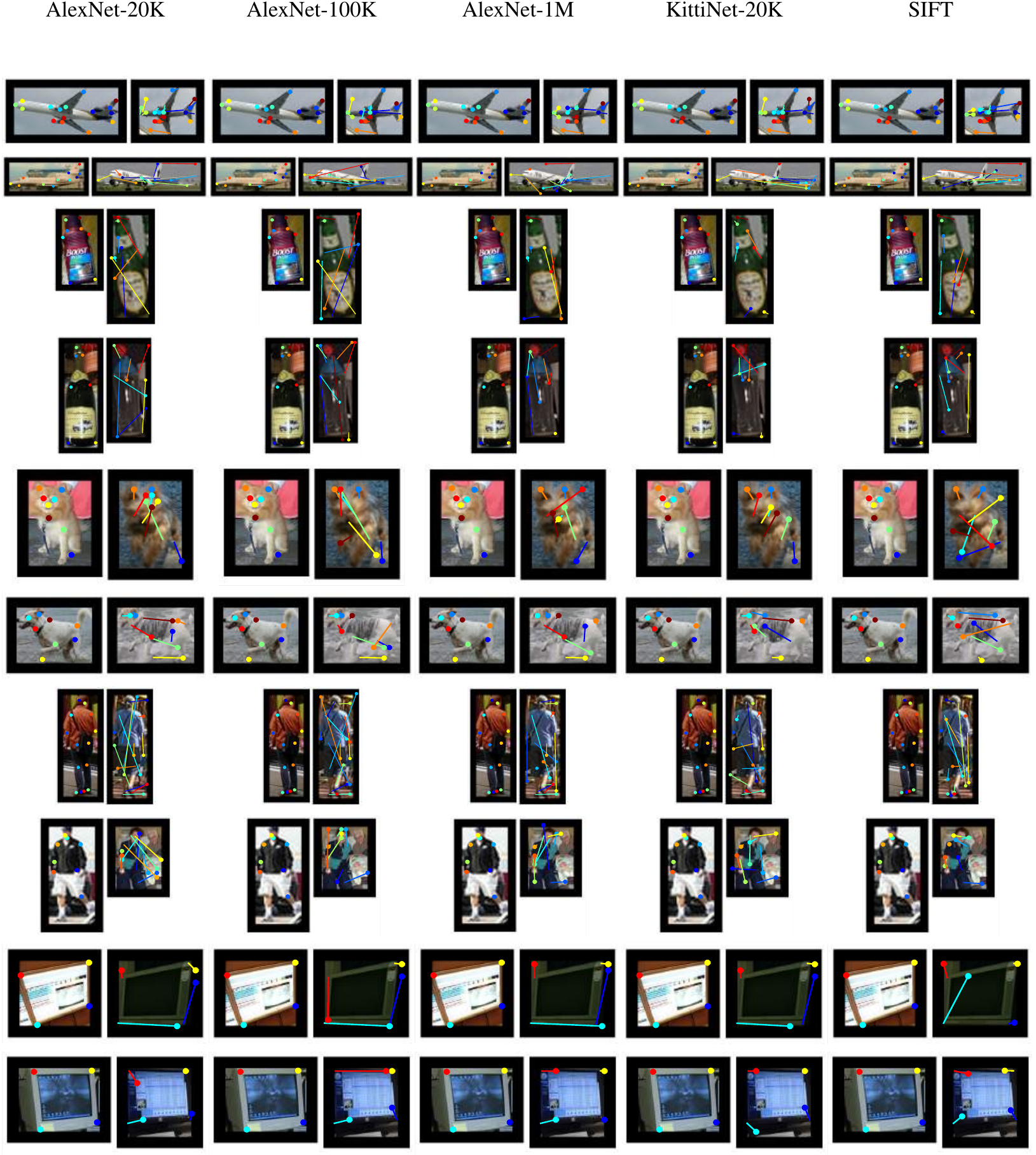}
 \end{center}
\caption{\label{fig:corresp} Example matchings between pairs of objects (randomly chosen) with viewpoints within $60$ degrees of each other, for classes ''aeroplane", "bottle", "dog", "person" and "tvmonitor" from PASCAL VOC. The matchings have been shown for features from layer conv-4 of AlexNet-20K, AlexNet-100K, AlexNet-1M, KittiNet-20K and SIFT. The left image shows the ground truth keypoints that were matched with the keypoints in the right image. Right images shows the location of the ground truth keypoint (shown by solid dot) and lines joining the predicted keypoint location (tip of the line) with the ground keypoint location. Please see section \ref{sec:key} for details of keypoint matching procedure and figure 4 in the main paper for numerical results. This figure is best seen in color and with zoom.}
\end{figure*}

Consider images of two instances of the same object class (for example airplane images as shown in first row of figure \ref{fig:corresp}) for which keypoint matching score needs to be computed. 

The images are pre-processed in the following way:
\begin{itemize}
\item Crop the groundtruth bounding box from the image. 
\item Pad the images by 30 pixels along each dimension. 
\item Resize each image so that the smallest side is 227 pixels. The aspect ratio of the image is preserved. 
\end{itemize}

\subsection{Keypoint Matching using CNN}
Assume that the $l^{th}$ layer of the CNN is used for feature computation. The feature map produced by the $l^{th}$ layer is of dimensionality $I\times J \times M$, where $(I,J)$ are the spatial dimensions and M is the number of filters in the $l^{th}$ layer. Thus, the $l^{th}$ layer produces a $M$ dimensional feature vector for each of the $I\times J$ grid position in the feature map. 

The coordinates of the keypoints are provided in the image coordinate system \cite{bourdev2010detecting}. For the keypoints in the first image, we first determine their grid position in the $I \times J$ feature map. Each grid position has an associated receptive field in the image. The keypoints are assigned to the grid positions for which the center of receptive field is closest to the keypoints. This means that each keypoint is assigned one location in the feature map. 

Let the $M$ dimensional feature vector associated with the $k^{th}$ keypoint in the first image be $F_1^k$. Let the $M$ dimensional feature vector at grid location $C_{ij}$ for the second image be $F_2(C_{ij})$. The location of matching keypoint in the second image is determined by solving:
\begin{equation}
C_{*} = argmin_{C_{ij}} \| F_1^{k} - F_2(C_{ij}) \|_2 
\end{equation}
$C_{*}$ is transformed into the image coordinate system by computing the center of receptive field (in the image) associated with this grid position. Let this transformed coordinates be $C_{*}^{im}$ and the coordinates of the corresponding keypoint (in the second image) be $C_{gt}^{im}$. The matching error for the $k^{th}$ keypoint ($E_k$) is defined as:
\begin{equation}
E_k =  \frac{\|C_{*}^{im} - C_{gt}^{im} \|_2}{L^2_{D}} 
\end{equation}
where, $L^2_{D}$ is the length of diagonal (in pixels) of the second image. As different images have different sizes, dividing by $L^2_D$ normalizes for the difference in sizes. 
The matching error for a pair of images of instances belonging to the same class is calculated as:
\begin{equation}
E_{instance} =  \frac{\sum_{k=1}^{K} E_k}{K}
\end{equation}
    
The average matching error across all pairs of the instance of the same class is given by $E_{class}$:
\begin{equation}
E_{class} =  \frac{\sum_{instance} E_{instance}}{\# pairs}
\end{equation}
where, $\# pairs$ is the number of pairs of object instances belonging to the same class. In Figure 4 of the main paper we report the matching error averaged across all the 20 classes.

\subsection{Keypoint Matching using SIFT}
SIFT features are extracted using a square window of size 72 pixels and a stride of 8 pixels using the open source code from \cite{vedaldi2010vlfeat}. The stride of 8 pixels was chosen to have a fair comparison with the CNN features. The CNN features were computed with a stride of 8 for layer conv-2 and stride of 16 for layers conv-3, conv-4 and conv-5 respectively. The matching error using SIFT was calculated in the same way as for the CNNs.  

\subsection{Effect of Viewpoint on Keypoint Matching}
Intuitively, matching instances of the same object that are related by a large transformation (i.e. viewpoint distance) should be harder than matching instances with a small viewpoint distance. Therefore, in order to obtain a holistic understanding of the accuracy of features in performing keypoint matching it is instructive to study the accuracy of matching as a function of viewpoint distance. 

\cite{vicente2014reconstructing} aligned instances of the same class (from PASCAL-VOC-2012) in a global coordinate system and provide a rotation matrix ($R$) for each instance in the class. To measure the viewpoint distance, we computed the riemannian metric on the manifold of rotation matrices $||log(R_i R_j^T)||_F$, where $log$ is the matrix logarithm, $||.||_F$ is the Frobenius norm of the matrix and $R_i, R_j$ are the rotation matrices for the $i^{th},j^{th}$ instances respectively. We binned the distances into 10 uniform bins (of 18\degree each). In Figure 4 of the main paper we show the mean error in keypoint matching in each of these viewpoints bin. The matching error in the $k^{th}$ bin is calculated by considering all the instances with a viewpoint distance $\leq k \times 18 \degree$, for $k \in [1,10]$. As expected we find that keypoint matching is worse for larger viewpoint distances.

\subsection{Matching Error for layers 2 and 5}

 \begin{figure*}[t]
       \begin{center}
 \includegraphics[width=0.49\linewidth]{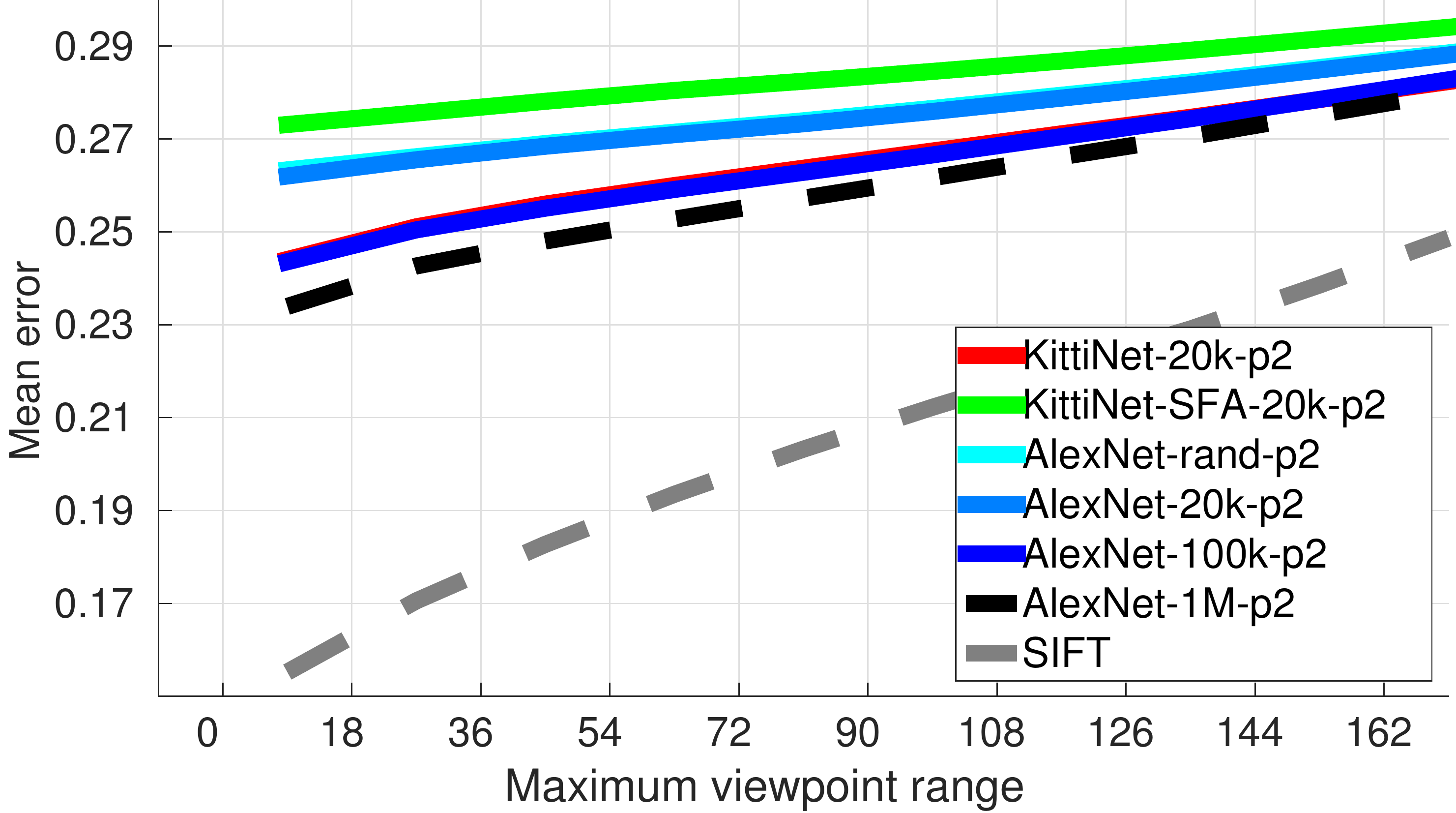}
 \includegraphics[width=0.49\linewidth]{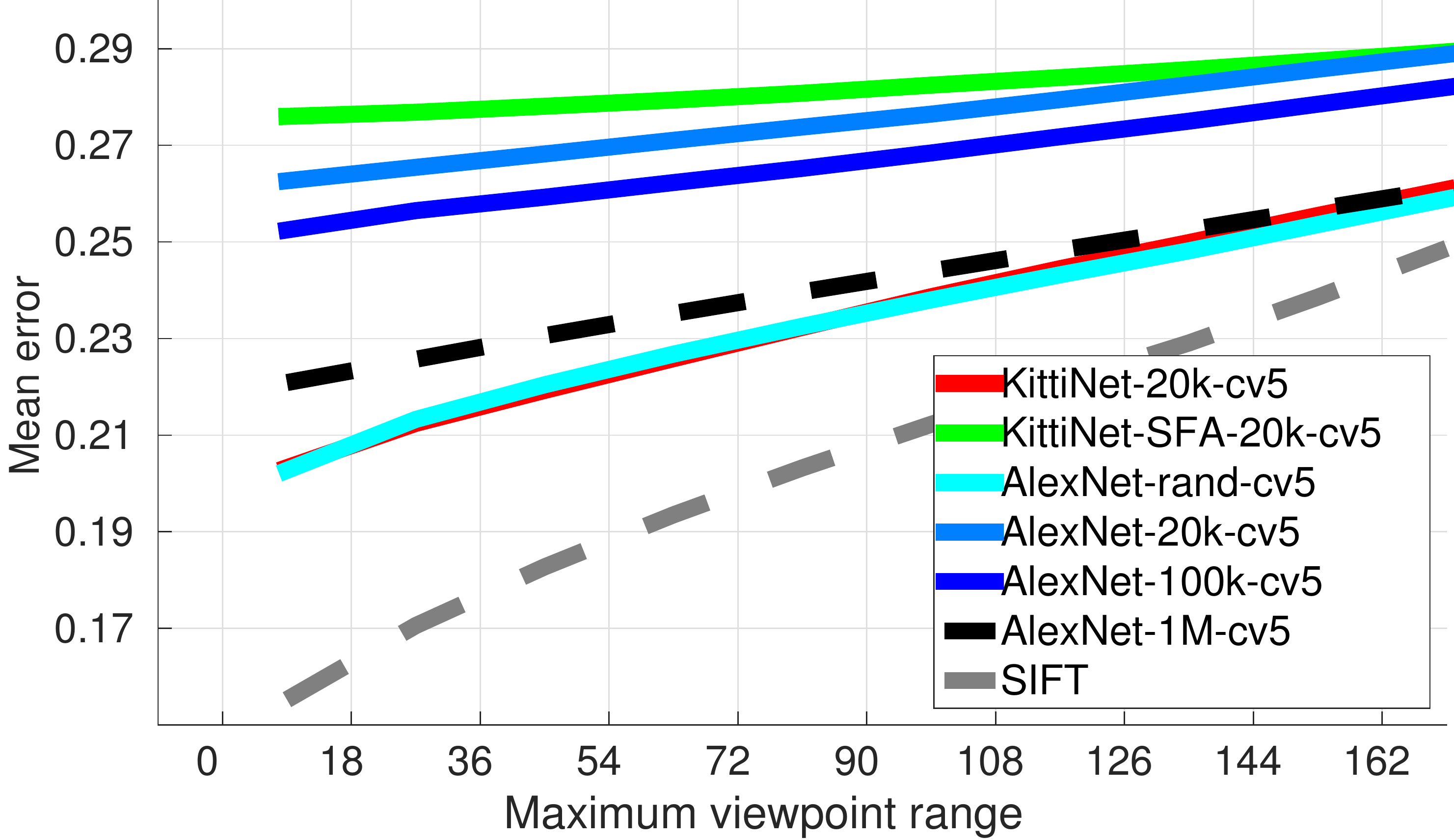}
       \end{center}
       \caption{Intra-class keypoint matching error as a function of viewpoint distance averaged over 20 PASCAL objects using features extracted from layers pool-2 (left) and conv-5 (right) of various networks used in this work. Please see section \ref{sub:keypoint} for more details.}
    \label{fig:alignmnent_err_sup}
\end{figure*}

\end{appendices}

{\small
\bibliographystyle{ieee}
\bibliography{all_refs}
}
    
\end{document}